\documentclass[conference]{IEEEtran}
\usepackage{lipsum}
\usepackage{graphicx}
\usepackage{hyperref}
\usepackage{fancyhdr}
\usepackage[numbers,sort,compress]{natbib}
\hypersetup{hidelinks,
	colorlinks=true,
	allcolors=black,
	pdfstartview=Fit,
	breaklinks=true
}
\IEEEoverridecommandlockouts

\usepackage[T1]{fontenc}

\begin{document}
 \title{SAR-UNet: Small Attention Residual UNet for Explainable Nowcasting Tasks}

\author{
    \IEEEauthorblockN{
        Mathieu Renault\IEEEauthorrefmark{2}, Siamak Mehrkanoon \IEEEauthorrefmark{2}\IEEEauthorrefmark{3}\IEEEauthorrefmark{1}\thanks{*Corresponding author.} 
    }
    \IEEEauthorblockA{\IEEEauthorrefmark{2} 
    Department of Advanced Computing Sciences, Maastricht University, Maastricht, Netherlands}
    \IEEEauthorblockA{ \IEEEauthorrefmark{3} Department of Information and Computing Sciences, Utrecht University, Utrecht, Netherlands}
    mathieu.renault@student.maastrichtuniversity.nl, 
    s.mehrkanoon@uu.nl
}





\maketitle
\renewcommand{\headrulewidth}{0pt}
\begin{abstract}
The accuracy and explainability of data-driven nowcasting models are of great importance in many socio-economic sectors reliant on weather-dependent decision making. This paper proposes a novel architecture called Small Attention Residual UNet (SAR-UNet) for precipitation and cloud cover nowcasting. Here, SmaAt-UNet is used as a core model and is further equipped with residual connections, parallel to the depthwise separable convolutions. The proposed SAR-UNet model is evaluated on two datasets, i.e., Dutch precipitation maps ranging from 2016 to 2019 and French cloud cover binary images from 2017 to 2018. The obtained results show that SAR-UNet outperforms other examined models in precipitation nowcasting from 30 to 180 minutes in the future as well as cloud cover nowcasting in the next 90 minutes. Furthermore, we provide additional insights on the nowcasts made by our proposed model using Grad-CAM, a visual explanation technique, which is employed on different levels of the encoder and decoder paths of the SAR-UNet model and produces heatmaps highlighting the critical regions in the input image as well as intermediate representations to the precipitation. The heatmaps generated by Grad-CAM reveal the interactions between the residual connections and the depthwise separable convolutions inside of the multiple depthwise separable blocks placed throughout the network architecture.

\end{abstract}
\begin{IEEEkeywords}
UNet, Precipitation Nowcasting, Cloud Cover Nowcasting, Deep Learning
\end{IEEEkeywords}

\IEEEpeerreviewmaketitle


\section{Introduction}

An accurate precipitation nowcast is essential in many domains such as urbanization, agriculture and tourism. Sudden heavy rainfalls can lead to major catastrophes such as floods and landslides and have a tremendous economic impact. Early Warning Systems (EWSs) use the nowcast/forecast to limit damages due to climate hazards. Therefore, quick, accurate and trustworthy forecasts are invaluable, especially for the near future, i.e., nowcasting. For instance, highly competitive mechanical sports such as Formula 1 rely on forecasts as detailed as minute-by-minute predictions on different parts of a circuit to decide the race setup and strategy. The aviation sector depends on various factors such as precipitation, wind speed and direction to ensure a safe journey for the passengers daily. State-of-the-art NWP based models rely on numerical methods simulating the intrinsic physical dynamics of the climate. These simulations are often complex and require vast computational resources. In this context, the development of advanced data-driven models such as Deep Learning (DL) in the past years has gained a lot of attention even has challenged the performance of NWP models \cite{c2}. 
Deep Learning based models have been previously successfully applied for forecasting tasks in a range of application domains including crop yield \cite{c4}, solar irradiance \cite{c6}, traffic \cite{c7} and weather 
\cite{c34, c35, c21, c27, c29, c30, c31, c32}.

The weather datasets are time-series with observations of a weather element at each time step. Recurrent Neural Networks (RNNs) are previously designed to include the order and time-dependency observed in datasets \cite{c8}. Over the last decade, literature has observed the development of Convolutional Neural Networks for several computer vision tasks \cite{c3}. In particular, LeNet-5 \cite{c9}, GoogLeNet \cite{c10} and U-Net \cite{c11} have shown promising results for image segmentation or object detection, especially in the medical field \cite{c12}\cite{c13}.
The literature on traditional computer vision tasks such as image classification is rich. However, image-to-image forecasting is rather a new field, and many architectures have yet to be developed to address the problems as the one studied here.


Despite outperforming performance of deep learning approaches, they are not yet fully transparent in how they reach to a particular decision. Recently, researchers have developed Explainable AI (XAI) techniques to help the user understand the logic that leads the model to the prediction \cite{c16}. Standard methods to achieve this are gradient-based saliency maps, Class Activation Maps, or Excitation Backpropagation. These post hoc methods produce heatmaps by computing the layers' activation after the network is trained. Other popular post hoc algorithms are perturbation-based, where one determines the essential features by altering or removing parts of the input and quantifying the impact on the prediction performance. 
Here, we use post hoc algorithms used previously in image classification and segmentation to give additional insights on model predictions. This paper proposes a novel architecture called Small Attention Residual UNet (SAR-UNet) which uses SmaAt-UNet \cite{c14} as core model and equips it with a Residual Connection parallel to each Depthwise Separable Convolution on both encoder and decoder paths. We evaluate the introduced model on precipitation and cloud cover nowcasting over the Netherlands and France respectively. The main contributions of the paper are the following:
\begin{itemize}
    \item We introduce a novel deep architecture for precipitation as well as cloud cover nowcasting. This network relies on the use of Depthwise Separable Convolution (DSC) with Residual connections and Attention mechanisms in its encoder and decoder. The SAR-UNet outperforms its predecessor, the SmaAt-UNet, in precipitation and cloud cover nowcasting tasks.
    \item We utilize Grad-CAM, a visual explanation technique, in order to provide additional insights on the nowcasts. Given an input image, Grad-CAM produces heatmaps of the activations in different levels of our network. To the best of our knowledge, this is the first adaptation of Grad-CAM for an image-to-image nowcasting task.
\end{itemize}

This paper is organized as follows. A brief overview of the related research works is given in Section II. Section III, introduces the proposed SAR-UNet model and the explains the used visual explanation technique. The experimental settings and description of the used datasets are given in Section IV. The obtained results are discussed in Section V and the conclusion is drawn in Section VI.

\section{Related Work}
Data-driven weather forecasting has recently gained a lot of attention. Among many successful deep learning architectures, Recurrent Neural Networks \cite{c28} and Long short-term memory (LSTM) \cite{c18} have shown promising results in sequential data analysis. Xu et al. \cite{c19} combined a Generative Adversarial Network with LSTM architecture to produce a network suitable for prediction using satellite cloud maps. An attempt to design an LSTM block that would encompass the treatment of the spatial aspect of the given input image led to the Convolutional LSTM Network (ConvLSTM) \cite{c20}, which has been successfully applied on precipitation nowcasting and outperformed the Fully Connected LSTM network.

UNet \cite{c11}, the widely used architecture in computer vision, has been previously extended and modified in various ways. For instance, the authors in \cite{c14} introduced the SmaAt-UNet, using a UNet core model, by replacing all regular convolutions with Depthwise-Separable Convolution (DSC) \cite{c17} as well as adding Convolutional Block Attention Modules (CBAMs) \cite{c26} to the encoder part. The explored DSC mechanism in the SmaAt-UNet significantly reduced the number of trainable parameters of the network. Moreover, the utilized CBAM blocks create attention maps both for the channel dimensions and the spatial dimensions, therefore can better learn the inter-channel as well as inter-spatial relationships of the weather variables. Diakogiannisa et al. \cite{c22} proposed a UNet backbone enhanced with residual connections parallel to the convolutional blocks. The use of the residual connections avoids the problem of exploding and vanishing gradients, allowing the creation of deeper networks. 

Explaining and interpreting deep weather forecasting models is still in its early stage. Abdellaoui et al. \cite{c23} used occlusion analysis to infer the importance of weather variables. The great advantage of this process is that it can be applied to any network architecture. In \cite{c21}, the author quantifies the uncertainty in the prediction through Test Time Dropout, a method that approximate the Bayesian inference of Bayesian Neural Networks. An uncertainty map is then produced to visualize and interpret the prediction. The authors in  \cite{c36, c37} use Local Interpretable Model-agnostic Explanations (LIME) in a weather forecast context to interpret the decisions from their model. LIME approximates the deep learning network with a simple model as a linear one to understand the relationships in the weather features. Most of the explainable AI techniques adapted to CNN architectures have been applied to image classification or segmentation tasks. For instance, the authors in \cite{c24} introduced Gradient-weighted Class Activation Maps (Grad- CAM) to interpret a CNN's output. 

\section{Method}

\subsection{Proposed SAR-UNet model}

\subsubsection{Architecture}
\begin{figure*}[htbp]
\centering
\includegraphics[scale=0.12]{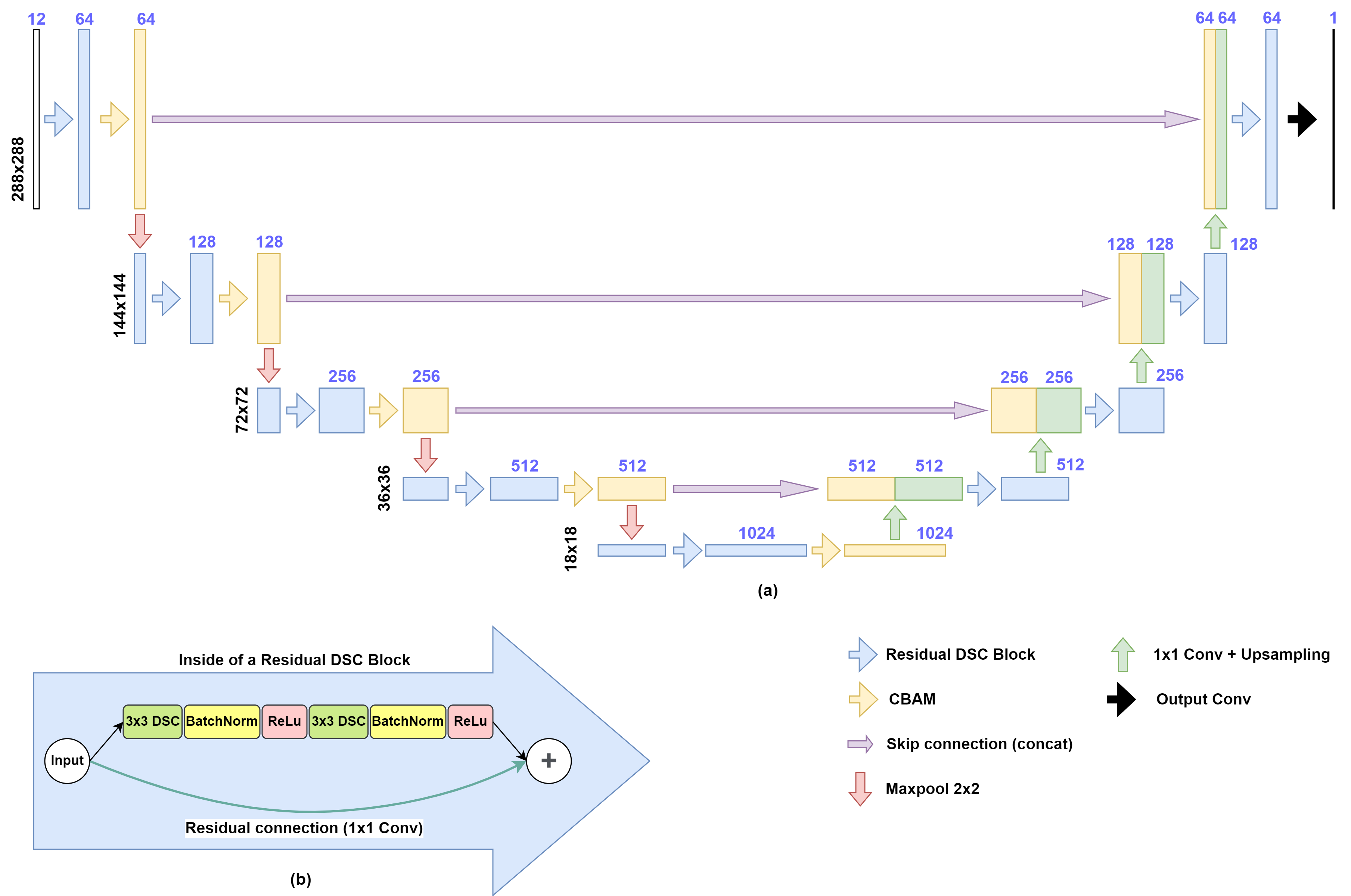}
\caption{SAR-UNet architecture. Arrows stand for a given transformation applied to the data, while the rectangles show the shape of the data after each step. The height and width of the images are shown on the left side at each level of the network. The number of filters used in the convolutions can be seen through the number of channels in the resulting images at each step. In the bottom left corner, we show the inside of a Residual DSC Block: the input goes, in parallel, through two DSCs on one side, through 1x1 convolution on the other, and both sides are then summed. The network uses 12 channels as input, corresponding to 60 minutes of image data.}
\label{SAR}
\end{figure*}

Fig. \ref{SAR}(a) shows a diagram that summarizes the proposed model. The model receives several images of size 288 by 288 stacked over the channels as input. Similar to UNet, its encoder-decoder transforms the data into spatially smaller images with more channels, then back into spatially larger images with fewer channels on each level before the output. Each level of the encoder is made of three transformations: Residual DSC Block (blue arrow), CBAM (yellow arrow), and 2x2 Max Pooling (red arrow). The CBAM's output is used as input for Max Pooling layer as well as it is send to the corresponding level (purple arrow) of the decoder part, where it is concatenated to the images from the lower level. The 2x2 Max Pooling reduces the spatial size of the images by 2, and its output is used as input to the next (lower) level of the encoder. A decoder level begins with 1x1 convolution followed by upsampling, represented by the green arrow in the diagram. The 1x1 convolution aims to halve the number of channels, while the upsampling doubles the spatial size of the images. The images are then concatenated to the CBAM's output over the channels and fed into a Residual DSC block. Note that the Residual DSC blocks double the number of channels in the encoder whilst they halve them in the decoder. A 1x1 output convolution (black arrow) layer is used in the end to obtain a single output image with the prediction.

\subsubsection{Residual DSC Block}
Residual connections act as a shortcut in the network and therefore make it easier to optimize the network parameters which itself can result in improving the training of the network and its accuracy. In addition, the residual connections allow for deeper networks to be trained without suffering from the vanishing gradients problem.
In the proposed Residual DSC Block, shown in Fig. \ref{SAR}(b), we combine the Residual connection with the Depthwise Separable Convolution mechanism.
The input goes through DSC with 3x3 kernel, Batch Normalization and ReLU twice. Moreover, this same input goes, in parallel, through a residual connection with 1x1 convolution to match the output channels of DSC. The output of the Residual DSC block is obtained by summing the output of the residual and the DSC paths.


\subsubsection{CBAM}
\begin{figure}[htbp]
\centering
\includegraphics[scale=0.18]{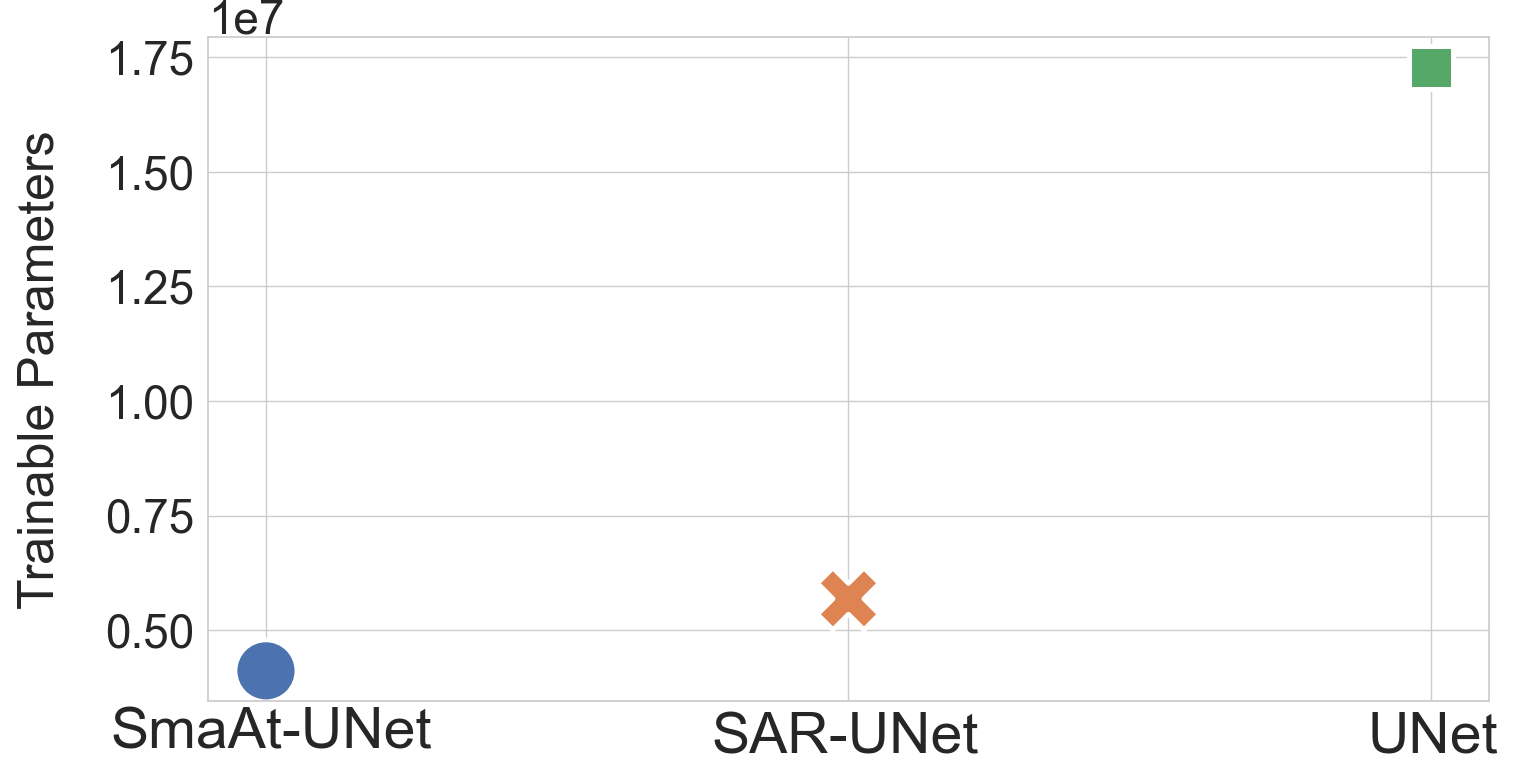}
\caption{Comparison of the number of trainable parameters between networks}
\label{params}
\end{figure}

We place CBAMs after each Residual DSC block in the encoder. As opposed to \cite{c14}, where the CBAM's output is only used in the skip connection, here it is also used as the input of the next level. The last adaptation of the SmaAt-UNet resides in the number of channels in the bottleneck layer. Our network has 1024 channels, twice as many as in the bottleneck of the SmaAt-UNet. For this reason, we use 1x1 Convolutions before each upsampling operation to reduce the channels in the next decoder level. These adaptations induce a slight increase in the network's total number of trainable parameters. We compare the total number of parameters of UNet, SmaAt-UNet and SAR-UNet in Fig. \ref{params}.

\subsection{Training} For training we followed the guidelines of \cite{c14}. The initial learning rate is set to 0.001 with a learning rate scheduler dividing the learning rate by 10 every time the validation loss has not decreased in four consecutive epochs. 
The Adam optimizer is used for training the model. The maximum number of epochs is set to 200, with an early stopping criterion of 15 epochs, effectively stopping training if there was no improvement over the last 15 epochs. The models are trained using a batch size of 6 on the Google Colab Pro platform with a GPU (Nvidia Tesla P100).

\subsection{Evaluation}
To evaluate the performance of the examined models, we use the Mean Squared Error (MSE) which is also used as the loss function during training. We also set a range of metrics after binarizing the output image. Following the lines of \cite{c14}, using a threshold of 0.5$mm/h$, we binarize each pixel of the output image and count the number of True Positives (TP), True Negatives (TN), False Positives (FP), and False Negatives (FN). Thanks to this, we get the Accuracy, Precision, Recall, and the F1-score of the model. We emphasize the MSE as an indicator of performance as this is a regression task, and the objective is to predict the exact quantity of precipitation in millimeters. Moreover, we compare the performance of the proposed model with that of SmaAt-UNet and the Persistence method, which uses the last input image of a sequence as the prediction image.

\subsection{Explainability}
Explainability is an essential part of deep learning nowcasting. 
Here, we explain the predictions through activation heatmaps. These heatmaps show which areas of the input are responsible for the high values in the output, thus making the nowcast more transparent. Grad-CAM was originally developed by \cite{c24} to produce a visual explanation of a classification network. It takes the gradient of the score for a class with respect to feature map activations to obtain neuron importance weights. Then it performs a linear combination of the forward activation maps followed by ReLU to obtain the final result. 
As mentioned before, after binarizing the output readar image for evaluation metrics, we obtain an image that is segmented with rain and no rain classes. We thus obtain explanations for our nowcasting task by transforming it into an image segmentation task. We set the algorithm to produce heatmaps for the rain class on the entire image.

\section{Experiments}
\subsection{Precipitation nowcasting}
We use the same precipitation map dataset as in \cite{c14, c27}. This dataset is collected from 2016 to 2019 by the Royal Netherlands Meteorological Institute (Koninklijk Nederlands Meteorologisch Instituut) with two radars in Den Helder and De Bilt, The Netherlands, capturing precipitation intensities every 5 minutes. Every pixel of the resulting image corresponds to one square kilometer of land, and its numerical value stands for the amount of rainfall, in hundredth of a millimeter. That means a pixel value of 1 translates to 0.01mm of rainfall. We follow the preprocessing steps of \cite{c14}, by cropping the images to a size of 288x288 pixels, centered on the Netherlands. As our interest is focused on images with rain, similar to \cite{c14}, we select images with a threshold set to 50\% of the total pixels having a value strictly higher than 0.

\begin{figure}[htbp]
\centering
\includegraphics[scale=0.13]{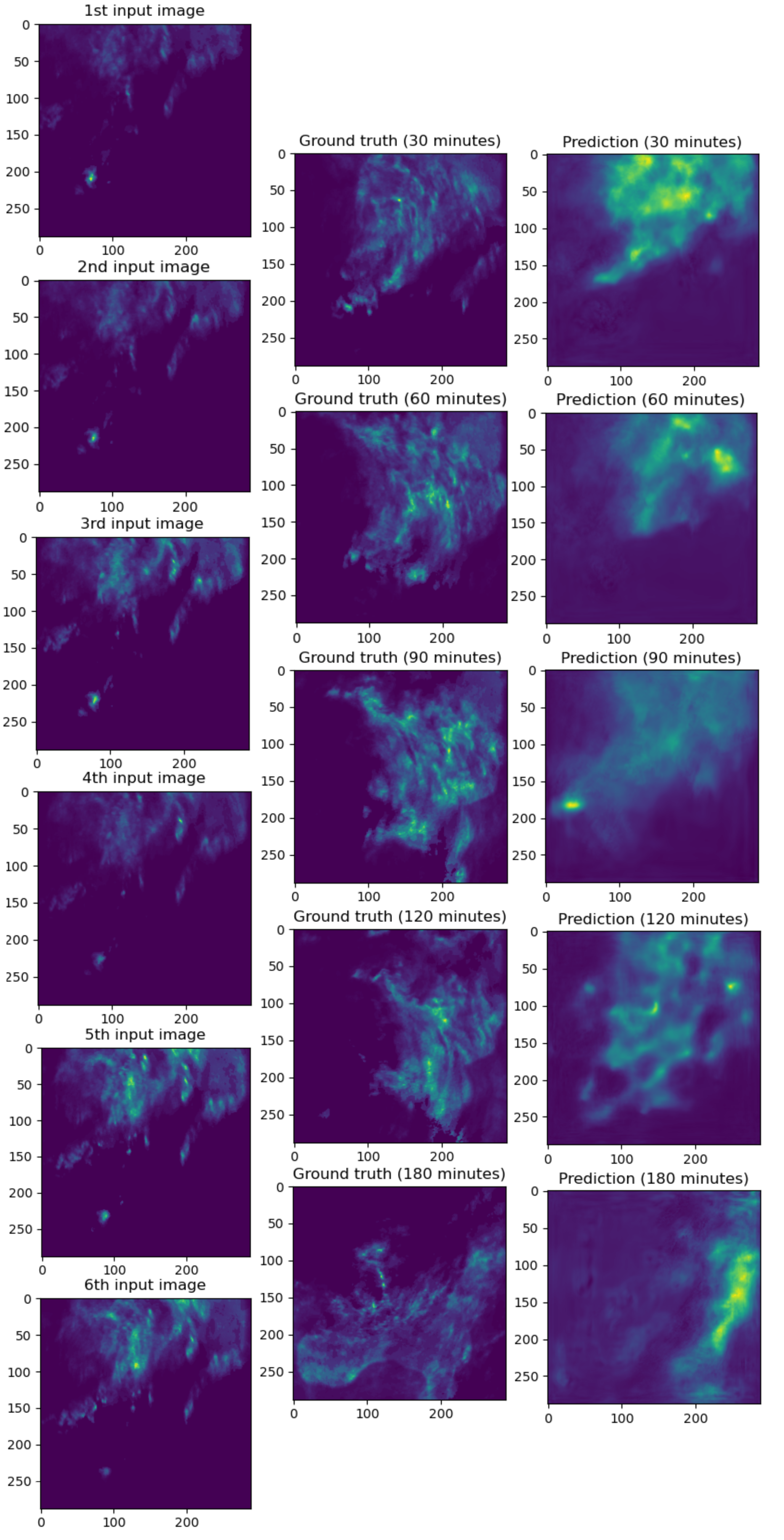}
\caption{Example of the nowcasting tasks, using 30 minutes input data. The first column presents the 6 input images. The second column shows the ground truth for nowcasting from 30 to 180 minutes ahead. The proposed model predictions are shown in the last column.}
\label{example pred}
\end{figure}

The sequence of images used as input for the network are stacked over the channel dimension. We have conducted a set of experiments to quantify the nowcasting performance in different situations. In particular, we use three different input sizes with 6, 12 and 18 channels, corresponding to 30, 60 and 90 minutes of data input for the network. For each of the input sizes, we conduct a range of nowcasting tasks: 30, 60, 90, 120 and 180 minutes ahead. This amounts to a total of 15 different nowcasting tasks. We used checkpoints to save the model after each epoch during training and for each setup we selected the checkpoint with lowest loss on the validation set. An example of the nowcastings obtained by our proposed SAR-UNet model is shown in Fig. \ref{example pred}. 

\subsection{Cloud cover nowcasting}
\begin{figure}[htbp]
\centering
\includegraphics[scale=0.35]{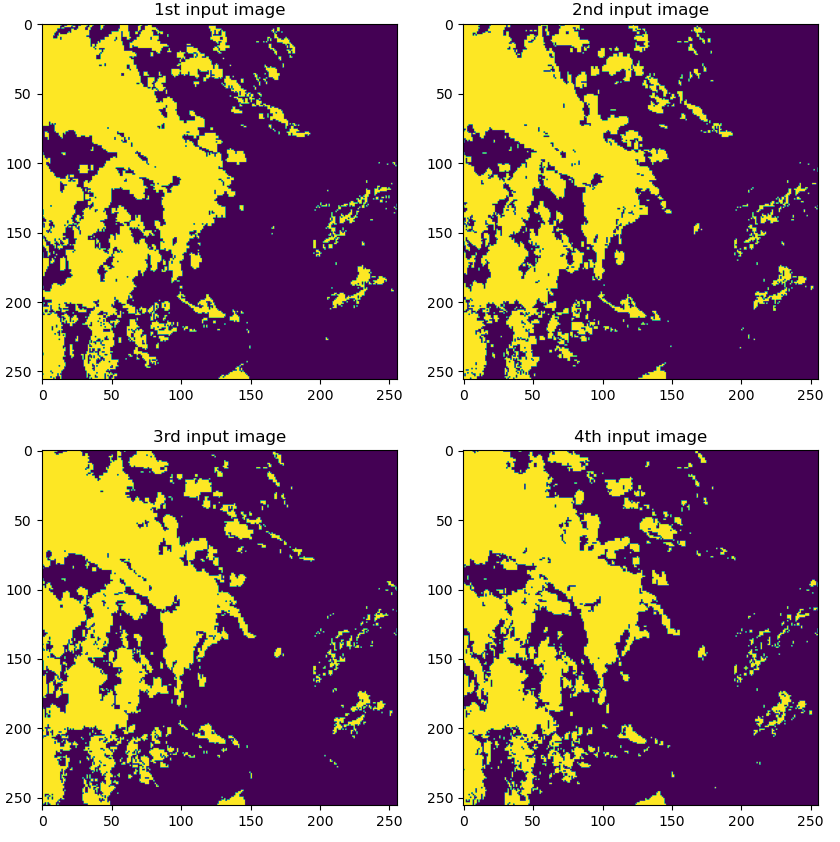}
\caption{Example of the cloud cover data input. The time interval between each image is 15 minutes.}
\label{example cloud}
\end{figure}

We also examine our model on the French cloud cover dataset used in \cite{c14}. The images in this dataset are binary and of size 256x256; each pixel has a value 1 if there is a cloud and 0 if there is none. The images are recorded every 15 minutes. Following the lines of \cite{c14}, we output six images (corresponding to the nowcasts for the next 1.5 hours ahead) using four images (corresponding to the past 1 hour data) as input. 
An example of the four input images is shown in Fig. \ref{example cloud}.

\section{Results and Discussion}
\subsection{Precipitation and cloud cover nowcasting}
\begin{figure}[htbp]
\centering
\includegraphics[scale=0.125]{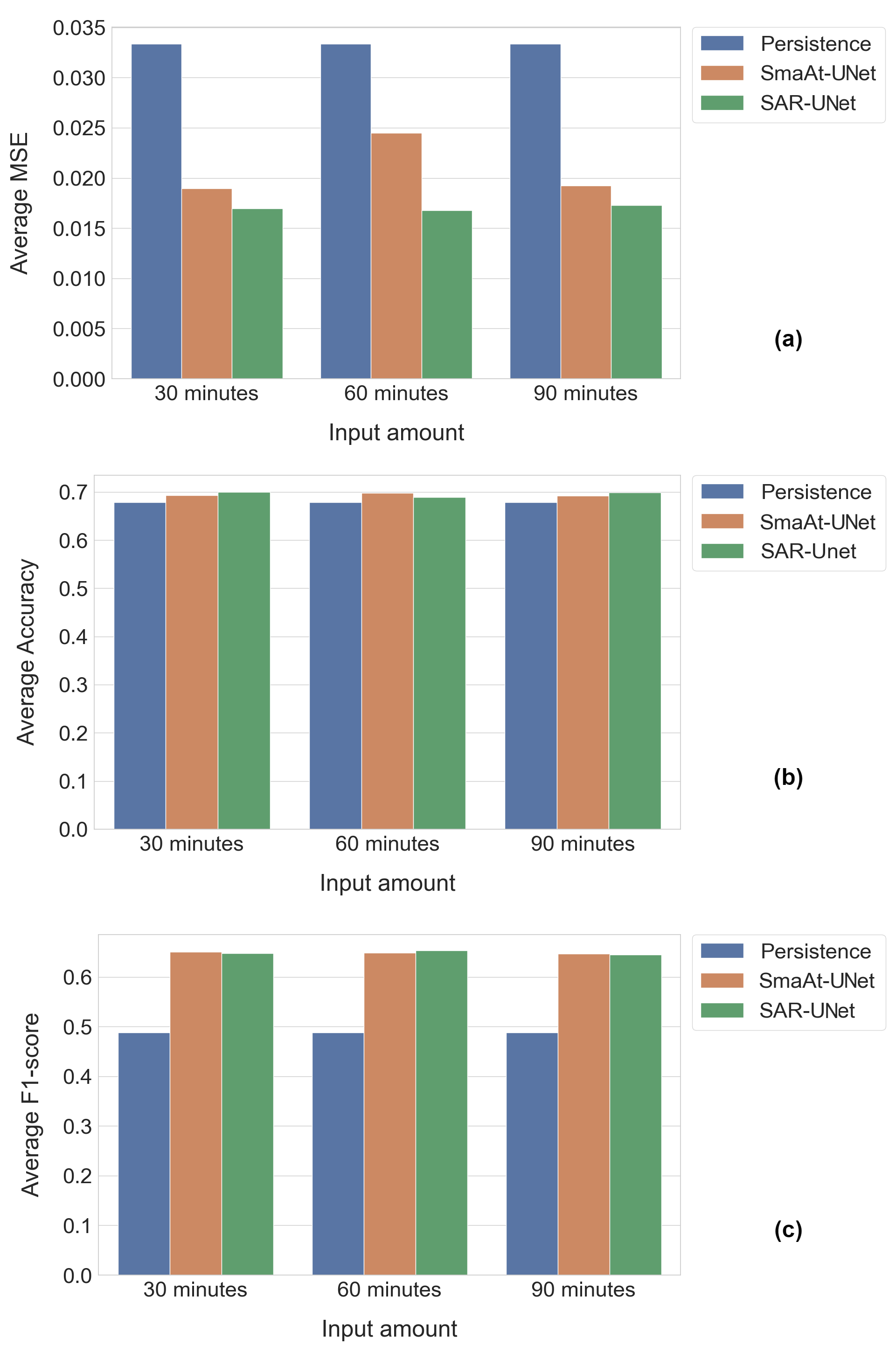}
\caption{The comparison of different metrics obtained by averaging over multi-step ahead nowcasts using different amounts of data as input. (a) The average MSE. (b) The average Accuracy. (c) The average F1-score.}
\label{barplot}
\end{figure}

This section presents the results obtained for the precipitation and cloud cover nowcasting tasks.
The obtained results of the precipitation task are summarized in Table \ref{results}, which show 
that SAR-UNet model is better than the other tested models in 13 of the 15 different setups when considering only the MSE.
\begin{table*}[htbp]
\caption{Comparison of the models for the \textbf{precipitation nowcasting tasks}.  $\uparrow$ means the best value for a metric is the highest while $\downarrow$ means the best value is the lowest.}
\begin{center}
\begin{tabular}{c|c|c c c c c c}
\hline
Input amount & Prediction time & Models & MSE $\downarrow$ & Precision $\uparrow$ & Recall $\uparrow$ & Accuracy $\uparrow$ & F1 score $\uparrow$  \\
\hline
& & Persistence &0,0249&	0,678	&0,643&	0,756	&0,66 \\
& 30 minutes ahead & SmaAt-UNet &0,0151&	0,675&	\textbf{0,883}&	0,8&	0,765 \\
& & SAR-UNet & \textbf{0,0124} & \textbf{0,73}&	0,835& \textbf{0,826}&	\textbf{0,779} \\
\cline{2-8}

& & Persistence &0,0318&0,603&	0,522	&0,698&	0,56 \\
&60 minutes ahead & SmaAt-UNet & \textbf{0,0160} &	\textbf{0,634}&	0,749&	\textbf{0,749}&	0,687\\
&& SAR-UNet & 0,0161&	0,607&	\textbf{0,799}&	0,736&	\textbf{0,69} \\
\cline{2-8}
 && Persistence &0,0360&	0,558	&0,43	&0,665&	0,486 \\
30 minutes & 90 minutes ahead & SmaAt-UNet &0,0177&\textbf{0,561}&	0,682&	\textbf{0,687}&	0,616\\
&& SAR-UNet & \textbf{0,0170} &0,532&\textbf{0,812}&0,668&\textbf{0,643}\\
\cline{2-8}
& & Persistence &0,0375&\textbf{0,532}&	0,355&	0,648&	0,426\\
&120 minutes ahead & SmaAt-UNet &0,0257&0,506	&\textbf{0,769}&	0,639&	\textbf{0,61}\\
&& SAR-UNet & \textbf{0,0189}	&0,524&	0,666&\textbf{0,654}&	0,586\\
\cline{2-8}
& & Persistence &0,0368&	0,478&	0,229	&\textbf{0,624}	&0,31\\
&180 minutes ahead & SmaAt-UNet & \textbf{0,0201} &	0,463&\textbf{0,75}&	0,588&\textbf{0,573}\\
&& SAR-UNet & 0,0203&\textbf{0,483} &	0,614&	0,616&	0,54\\
\hline

& & Persistence &0,0249&	0,678	&0,643&	0,756	&0,66 \\
&30 minutes ahead & SmaAt-UNet & 0,0248&	0,677&\textbf{0,878}	&0,801	&0,764
\\
&& SAR-UNet & \textbf{0,0120} &\textbf{0,697}&0,868&\textbf{0,813}&\textbf{0,774}\\
\cline{2-8}
& & Persistence &0,0318&0,603&	0,522	&0,698&	0,56 \\
&60 minutes ahead & SmaAt-UNet & 0,0166&	0,582	&\textbf{0,843}&	0,719&	\textbf{0,688}\\
&& SAR-UNet & \textbf{0,0163} &\textbf{0,623}&	0,747&\textbf{0,74}&	0,679\\
\cline{2-8}
& & Persistence &0,0360&\textbf{0,558}	&0,43	&0,665&	0,486\\
60 minutes &90 minutes ahead & SmaAt-UNet & 0,0314&	0,553&	0,737&\textbf{0,684}&\textbf{0,632}\\
&& SAR-UNet & \textbf{0,0185} &0,542&\textbf{0,747}&	0,674&	0,628 \\
\cline{2-8}
& & Persistence &0,0375&	0,532&	0,355&	0,648&	0,426\\
&120 minutes ahead & SmaAt-UNet & 0,0196& \textbf{0,54}&	0,653&	\textbf{0,667}&	0,591\\
&& SAR-UNet & \textbf{0,0176} & 0,485 & \textbf{0,831}&	0,613&\textbf{0,612} \\
\cline{2-8}
& & Persistence &0,0368&	0,478&	0,229	&\textbf{0,624}	&0,31\\
&180 minutes ahead & SmaAt-UNet & 0,0302&\textbf{0,488}&	0,685&	0,619&	0,57\\
&& SAR-UNet & \textbf{0,0196} & 0,477	&\textbf{0,712}&	0,607&\textbf{0,571}\\

\hline
& & Persistence &0,0249&	0,678	&0,643&	0,756	&0,66\\
&30 minutes ahead & SmaAt-UNet & 0,0131&\textbf{0,72}0&	0,84&\textbf{0,821}&	\textbf{0,776}\\
&& SAR-UNet & \textbf{0,0125} & 0,703&\textbf{0,851}&	0,813&	0,77 \\
\cline{2-8}
& & Persistence &0,0318&0,603&	0,522	&0,698&	0,56 \\
&60 minutes ahead & SmaAt-UNet & 0,0197&	0,576&	0,792&	0,709	&0,667\\
&& SAR-UNet & \textbf{0,0171} &\textbf{0,615}&\textbf{0,802}&\textbf{0,742}&\textbf{0,696}\\
\cline{2-8}
& & Persistence &0,0360&\textbf{0,558}	&0,43	&0,665&	0,486\\
90 minutes &90 minutes ahead & SmaAt-UNet & 0,0205&	0,547&\textbf{0,752}&\textbf{0,679}&\textbf{0,633}\\
&& SAR-UNet & \textbf{0,0176} &	0,539&	0,749&	0,672&	0,627 \\
\cline{2-8}
& & Persistence &0,0375&\textbf{0,532}&	0,355&	0,648&	0,426\\
&120 minutes ahead & SmaAt-UNet &0,0211	&0,499&\textbf{0,754}&	0,631&	\textbf{0,601}\\
&& SAR-UNet & \textbf{0,0191} &	0,52&	0,659&\textbf{0,65}&	0,581 \\
\cline{2-8}
& & Persistence &0,0368&	0,478&	0,229	&\textbf{0,624}&0,31\\
&180 minutes ahead & SmaAt-UNet & 0,0217&\textbf{0,487}&\textbf{0,649}&	0,619&\textbf{0,556}\\
&& SAR-UNet & \textbf{0,0202} &	0,484&	0,641&	0,616&	0,552 \\
\hline
\end{tabular}
\label{results}
\end{center}
\end{table*}
which 
We emphasize this by plotting the average MSE (average of all minutes ahead predictions) by the amount of input used for each model in Fig. \ref{barplot}. 
In this Figure, we compare the performance of the examined models and visualize the superiority of SAR-UNet model over the SmaAt-UNet and the Persistence models. In addition, there does not seem to be an substantial increase or decrease in the model performance when varying input data amounts. Table \ref{results} shows that nowcasting further away in the future is a more challenging task, and in general the losses increase as the minutes ahead increases. 
For Precision, Recall and F1-score, the SAR-UNet is better than the SmaAt-UNet in about as many setups as when the SmaAt-UNet is better than the SAR-UNet. Therefore, SAR-UNet and SmaAt-UNet are comparable in terms of these tested metrics. This result might be explained by the arbitrary choice of a threshold for binarization of the prediction used in calculating these metrics. The obtained metrics for the cloud cover nowcasting task are tabulated in Table \ref{results cloud cover}.
Similar to precipitation task, here SAR-UNet in most of the cases outperforms the SmaAt-UNet and Persistence models.
\begin{table}[h!]
\caption{Comparison of the models for the \textbf{cloud cover nowcasting}.  $\uparrow$ means the best value for a metric is the highest while $\downarrow$ means the best value is the lowest.}
\begin{center}
\resizebox{\columnwidth}{!}{%
\begin{tabular}{c c c c c c}
\hline
Models & MSE $\downarrow$ & Precision $\uparrow$ & Recall $\uparrow$ & Accuracy $\uparrow$ & F1 score $\uparrow$  \\
\hline
Persistence &0.1491 & 0.872 &0.872 & 0.851& 0.872\\
SmaAt-UNet & 0.0794 & \textbf{0.892} & 0.921 & 0.889 &0.906 \\ 
SAR-UNet & \textbf{0.0787} & \textbf{0.892} & \textbf{0.923} & \textbf{0.890} & \textbf{0.907}\\
\hline
\end{tabular}
}
\label{results cloud cover}
\end{center}
\end{table}

\subsection{Activation heatmaps}
\begin{figure}[h!]
\centering
\includegraphics[scale=0.19]{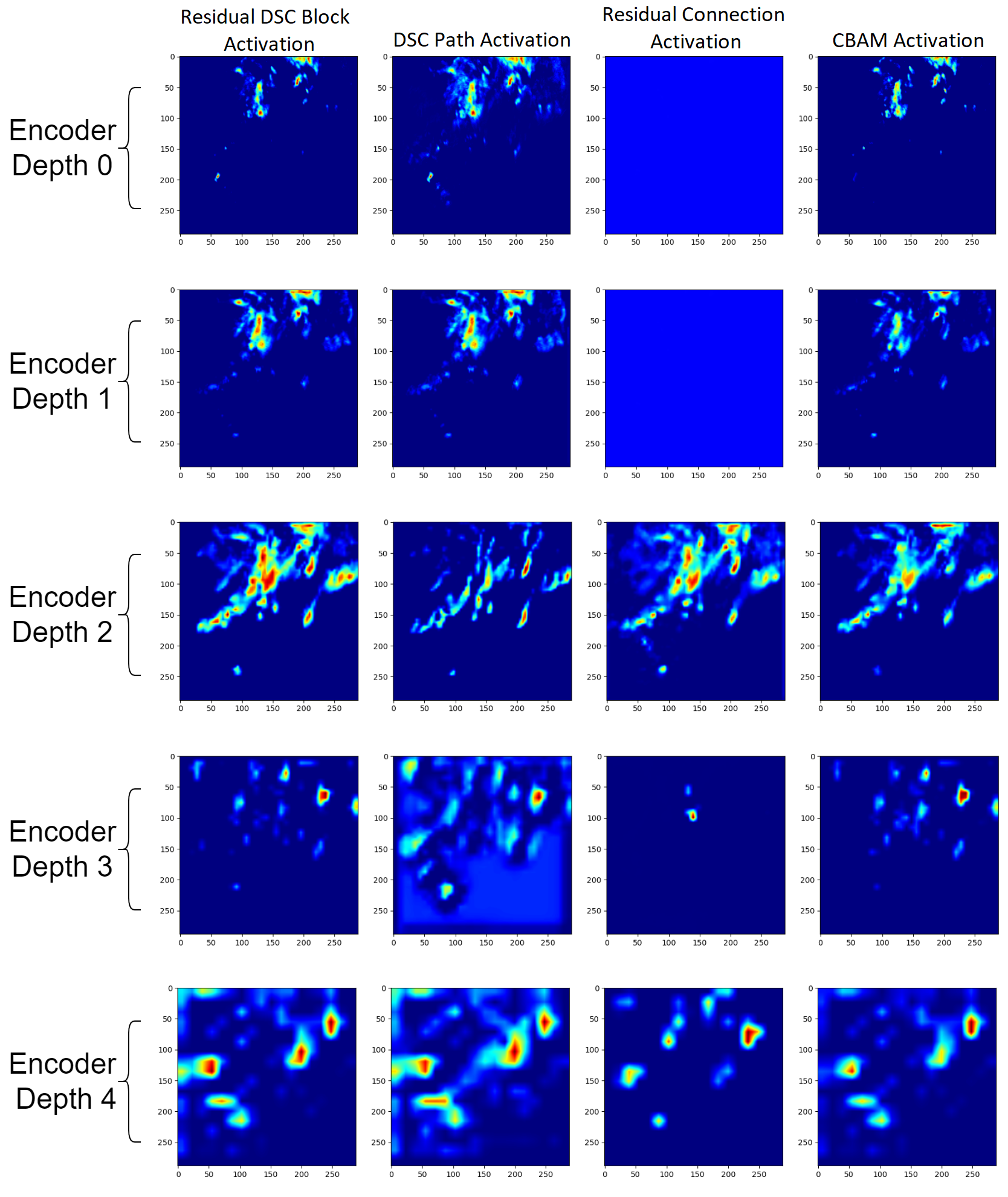}
\caption{Activation heatmaps obtained using Grad-CAM on all pixels predicted with rain after binarization. The first column represents the activation from an entire Residual DSC Block, while the second and third columns focus on parts of it, i.e., the activation of a DSC and the activation of the Residual connection respectively. The activations of the CBAM are shown on the last column. Each row stands for a level (depth) of the encoder part of SAR-UNet model. Thus, the first row is the first transformation for the input, while the bottom row is the last transformation of the encoder part.}
\label{expl}
\end{figure}

\begin{figure}[h!]
\centering
\includegraphics[scale=0.19]{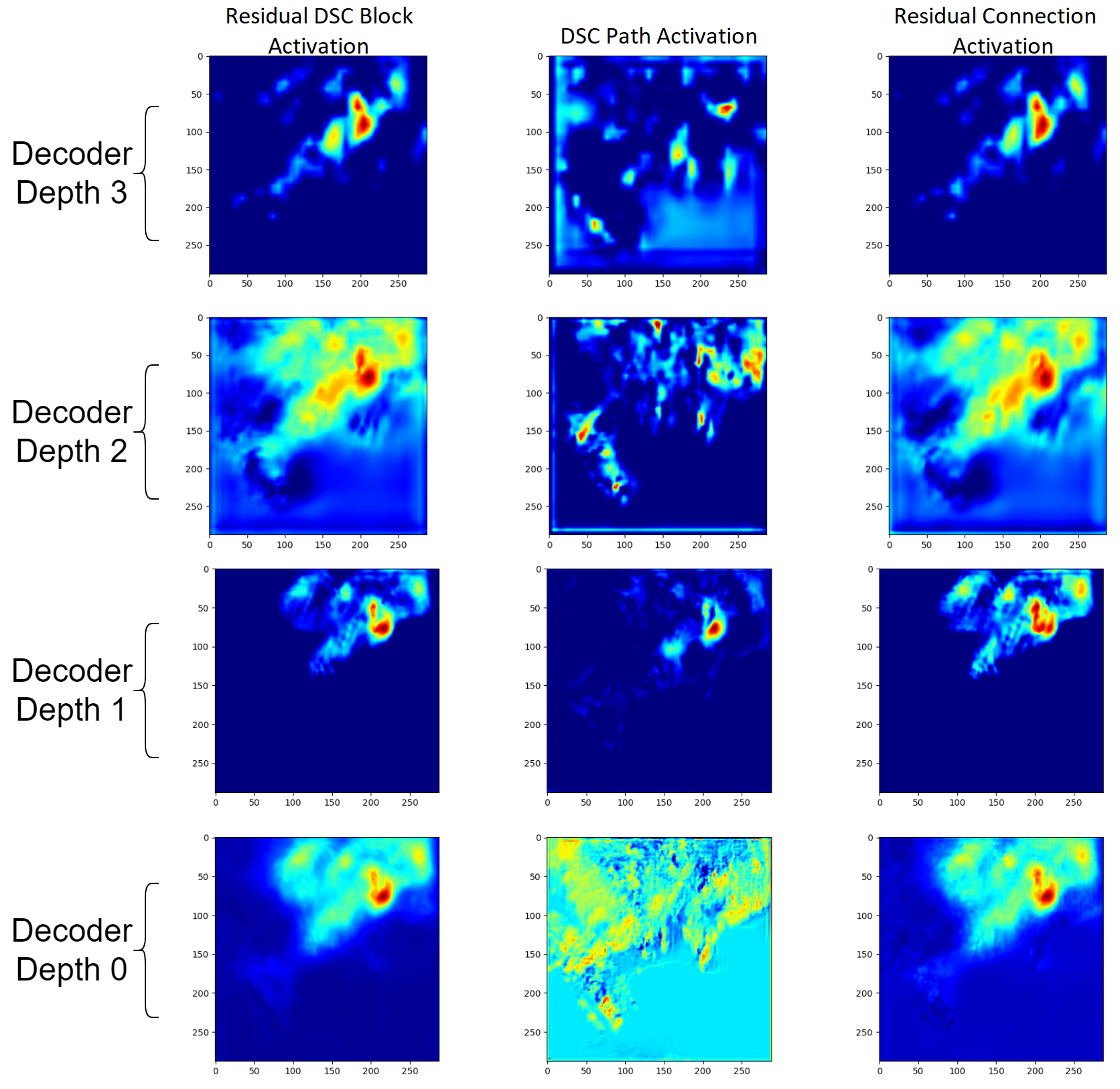}
\caption{Activation heatmaps obtained through the use of Grad-CAM on all pixels predicted with rain after binarization. The first column represents the activation from an entire Residual DSC Block, while the second and third columns focus on parts of it, i.e., the activation of a DSC and the activation of the Residual connection respectively. Each row stands for a level of the decoder part of SAR-UNet model. Thus, the first row is the bottom level of the decoder, while the final row is the last transformation before the output convolutional layer.}
\label{expl up}
\end{figure}

The collection of heatmaps obtained using Grad-CAM bring additional clarity to how SAR-UNet functions and what parts of the images are responsible for the prediction. In Fig. \ref{expl} and Fig. \ref{expl up} we show the heatmaps of network's activations for the precipitation nowcasting. These heatmaps are obtained using the same data shown in Fig. \ref{example pred}. The figures contain heatmaps of the activations of different inner layers of the SAR-UNet to explain how they operate. Therefore, we output a heatmap for the whole Residual DSC blocks, as well as each DSC sequence and Residual connection inside of these blocks, and finally for all CBAMs in the encoder path of the model. We split the plots between the encoder part in Fig. \ref{expl} and decoder part in Fig. \ref{expl up} where the complete path followed by the data, from the very beginning to the very end are shown. This provides additional insights on which part of the input image is most important at each layer and level of the model and that how these information are combined to give a final prediction.

We start by analyzing the first column of the figures, representing the Residual DSC Blocks. The first three levels of the encoder in Fig. \ref{expl} seem to be focused on the areas where the last input image presents the highest precipitation amounts. In particular, first only the areas with very high precipitation are activated in the encoder depth 0, progressing to a larger activated area in encoder depth 2 corresponding to areas with rain in the input. The deeper levels of the network, shown in Fig. \ref{expl} and \ref{expl up}, namely Encoder depth 3,4 and Decoder depth 3, appear to activate in more abstract areas, with a few dots being significantly activated, leaving the rest of the image inactivated. The final three levels (depth 2,1 and 0) of the decoder in Fig. \ref{expl up} are much more similar to the prediction made by the network. These layers smooth out the prediction so that areas vary gradually from little rain to substantial rain, giving a more realistic appearance to the prediction.

\begin{figure}[h]
\centering
\includegraphics[scale=0.07]{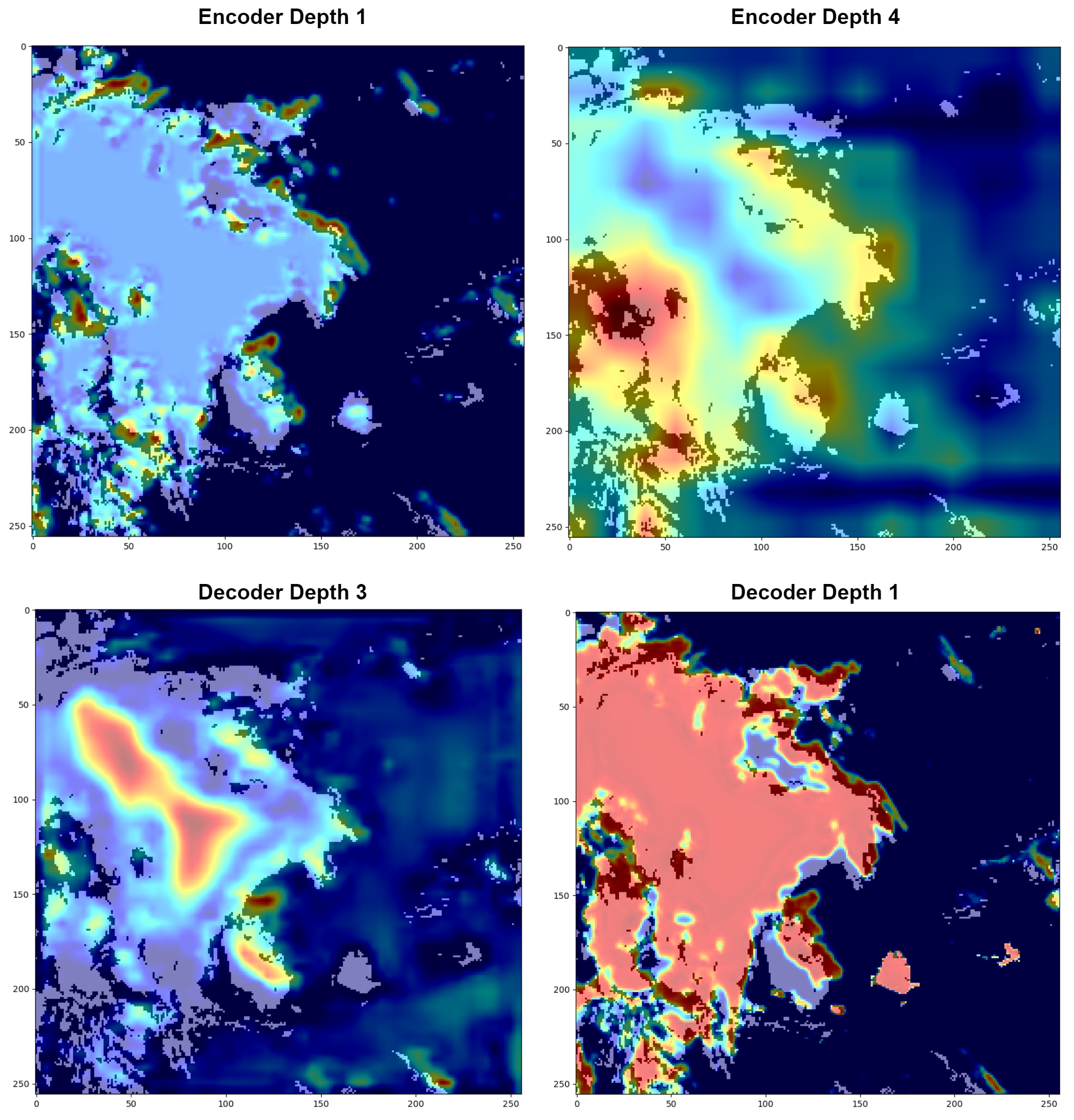}
\caption{The obtained activation heatmaps of four different residual DSC blocks within the SAR-UNet, for the cloud cover nowcasting task.}
\label{cloud cover expl}
\end{figure}

The activation from a Residual DSC block is generally similar to the ones from the DSC path and the Residual connection as it is the sum of both. Nevertheless, it appears in Fig. \ref{expl} that the DSC path is closer to the output of the Residual DSC block, while the Residual connection is less important with the first two depth levels being uniformly activated. On the other hand, in Fig. \ref{expl up} it is the Residual connection that is activated similarly to the Residual DSC Block of the same row. Therefore, in the encoder and decoder parts, we can infer that the two paths of a Residual DSC Block switch roles and their importance change gradually. Finally, the fourth column in Fig. \ref{expl} presents the activations of the CBAMs in the Encoder part of the network. We observe that the CBAMs activate almost exactly as the Residual DSC Blocks. The visible difference is in the intensity of the activation, especially near the points where the activation is the maximum, i.e., the red and orange areas. The activation of the CBAM is less intense in these areas, leading to fewer red areas and more yellow and green ones. Fig. \ref{cloud cover expl} shows the activation heatmaps of Encoder depth 1 and 4, and Decoder depth 3 and 1 of the SAR-UNet for the cloud cover dataset.
We notice the Encoder depth 1, placed at the beginning of the network, is activated at the borders between cloud and non-cloud zones with high precision. Encoder depth 4 is also activated on these borders but in wider patches, leading to larger red zones on the heatmap. In Decoder depth 2, we observe activation zones in the center of the cloudy areas of the image. It is focused on the inside of the cloud area delimited in the previous layers. Decoder depth 1 is one of the final layers of the network. Its activation heatmap is very different from the other shown heatmaps. This layer's activation covers almost entirely the cloud areas of the image. It is therefore a combination of the previous layers, with the borders and the centre of the clouds all activated areas.

\section{Conclusion}
In this paper a novel Small Attention Residual UNet (SAR-UNet) is proposed for weather element nowcasting tasks.
The model is based on a U-shape convolutional network that combines Depthwise Separable Convolutions, Residual Connections and the Convolutional Block Attention Modules to outperform its predecessor, the SmaAt-UNet.
The proposed model is evaluated on two nowcasting tasks, i.e., precipitation and cloud cover nowcasting. The experimental results demonstrate that SAR-UNet model outperforms the other tested models for the studied datasets.
In order to shed lights on the explainability of the inner workings of the proposed SAR-UNet model, we have visualized the activation heatmaps for the nowcasts using Grad-CAM technique. The implementation of our SAR-UNet model is available at GitHub\footnote{\href{https://github.com/mathieurenault1/SAR-UNet}{https://github.com/mathieurenault1/SAR-UNet}}.

\label{sect:c}

\bibliographystyle{IEEEtran}
\bibliography{main}
\end{document}